\documentclass[journal]{IEEEtran}

\usepackage{xcolor,soul,framed} %,caption

\colorlet{shadecolor}{yellow}
\usepackage[pdftex]{graphicx}
\graphicspath{{../pdf/}{../jpeg/}}
\DeclareGraphicsExtensions{.pdf,.jpeg,.png}

\usepackage[cmex10]{amsmath}
\usepackage{array}
\usepackage{mdwmath}
\usepackage{mdwtab}
\usepackage{eqparbox}
\usepackage{url}
\usepackage{caption}
\usepackage{siunitx}
\usepackage{float}
\usepackage{cite} 
\usepackage[square,sort]{natbib}
\usepackage{hyperref}  %hyperref still needs to be put at the end!
\usepackage{amssymb}
\usepackage{textcomp}
\begin{document}
\bstctlcite{IEEEexample:BSTcontrol}
    \title{Active Learning for Neural Machine Translation}
  \author{Neeraj~Vashistha,~\IEEEmembership{Queen Mary University of London,}
      Kriti~Singh,~\IEEEmembership{University of Bath}\\
      Ramakant~Shakya,~\IEEEmembership{DigilyticsAI, Gurugram, India}% <-this % stops a space

  \thanks{This document is the results of the research
   by individual contributors and is not funded by any foundation.}}
  % \thanks{M. Roberg is with TriQuint Semiconductor, 500 West Renner Road Richardson, TX 75080 USA (e-mail: michael.roberg@tqs.com).}% <-this % stops a space
%   \thanks{T. Reveyrand is with the XLIM Laboratory, UMR 7252, University of Limoges, 87060 Limoges, France (e-mail: tibault.reveyrand@xlim.fr).}%
%   \thanks{I. Ramos and Z. Popovic are with the Department of Electrical, Computer and Energy Engineering, University of Colorado, Boulder, CO, 80309-0425 USA (e-mail: ignacio.ramos@colorado.edu; zoya.popovic@colorado.edu).}% <-this % stops a space
%   \thanks{E. Falkenstein is with Qualcomm Inc., 6150 Lookout Road
% Boulder, CO 80301 USA (e-mail: erez.falkenstein@gmail.com).}}  

% \markboth{IEEE TRANSACTIONS ON MICROWAVE THEORY AND TECHNIQUES, VOL.~60, NO.~12, DECEMBER~2012
% }{Roberg \MakeLowercase{\textit{et al.}}: High-Efficiency Diode and Transistor Rectifiers}

% ====================================================================
\maketitle

% === ABSTRACT ====================================================================
% =================================================================================
\begin{abstract}
%\boldmath
The machine translation mechanism translates texts automatically between different natural languages, and Neural Machine Translation (NMT) has gained attention for its rational context analysis and fluent translation accuracy. However, processing low-resource languages that lack relevant training attributes like supervised data is a current challenge for Natural Language Processing (NLP). We incorporated a technique known Active Learning with the NMT toolkit Joey NMT to reach sufficient accuracy and robust predictions of low-resource language transla- tion. With active learning, a semi-supervised machine learning strategy, the training algorithm determines which unlabeled data would be the most beneficial for obtaining labels using selected query techniques. We implemented two model-driven acquisition functions for selecting the samples to be validated. This work uses transformer-based NMT systems; baseline model (BM), fully trained model (FTM) , active learning least confidence based model (ALLCM), and active learning margin sampling based model (ALMSM) when translating English to Hindi. The Bilingual Evaluation Understudy (BLEU) metric has been used to evaluate system results. The BLEU scores of BM, FTM, ALLCM and ALMSM systems are 16.26, 22.56 , 24.54, and 24.20, respectively. The findings in this paper demonstrate that active learning techniques helps the model to converge early and improve the overall quality of the translation system.
\end{abstract}

% === KEYWORDS ====================================================================
% =================================================================================

% \fi
\IEEEpeerreviewmaketitle

% ====================================================================
% ====================================================================
% ====================================================================

% === I. INTRODUCTION =============================================================
% =================================================================================
\section{Introduction}

One of the sub-field of computational linguistics known as Machine translation looks into the software utilization for speech or text language translation. \citep{Locke1955MachineTO}, began the advances in machine translation. Machine translation systems developed in a progression of systems from rule-based to corpus-based approaches. Corpora-based machine translation systems are classified into Example-Based Machine Translation (EBMT), Statistical Machine Translation (SMT) and Neural Machine Translation (NMT). The scope of EBMT is quite limited because, it requires a large corpus, and not everything can be covered by example, the spoken languages are too vivid, diverse and ambiguous. Hence, SMT came into existence which relies upon bayesian inference. SMT predicts translation probabilities of phrase pairs in corresponding source-target languages. By increasing the size of the dataset, the probability of a certain pair of phrases can be enhanced. However, the inability to achieve context information, differently trained components and system complexity are the weak points of SMT, which led to the development of the NMT system \citep{S2017StatisticalVR}. The NMT system can handles sequence-to-sequence learning problems for variable length source and target sentences and long-term dependency problems. The NMT system improves translation prediction and has excellent context-analyzing properties.

The superiority of NMT over phrase-based SMT is undeniable, and neural networks are used in most online machine translation engines. However, in spite of the growth achieved in the domain of machine translation, the idea of NMT system development being data-hungry continues to be a vital issue in expanding the work for any low-resource languages. In order to train a high-quality translation model, NMT requires a large bilingual corpus. However, creating parallel corpora for most low-resource language pairs is costly and requires human effort. Nevertheless, the language and geographical coverage of NMT have yet to hit new heights due to resource accessibility concerns and a preference for well-established assessment benchmarks. This compels additional research into the present state of NMT utilizing novel techniques for such languages, particularly at a point when an efficient and equal exchange of knowledge across borders and cultures is a pressing need.

In this research, we are utilising active learning, an iterative semi-supervised framework to reduce the cost of data acquisition for machine translation. Active learning provides solutions to two challenging problems of data, namely, its quantity and quality. In active learning, the learner is able to query an oracle for labelling new data points. The quantity of labelling required to understand a concept can be substantially lower than annotating the entire unlabelled dataset because the learner selects the data points for annotation \citep{BALCAN200978}. This approach is helpful in low-resource scenarios where unlabeled data is abundant, but manual labelling is expensive. Numerous NLP domains, including categorization, sequence labelling, spoken language comprehension, and machine translation, have benefited from the application of active learning. \citep{cohn1994active} \citep{guo2007optimistic} \citep{DAGAN1995150} \citep{settles2008analysis} \citep{tur2005combining} \citep{ambati2011multi} \citep{Eck2008_1000010682} \citep{peris-casacuberta-2018-active} \citep{8629116}.
\begin{table*}
    \centering
    \begin{tabular}{|l|l|l|l|l|l|l|l|}
    \hline
        \textbf{System} & \multicolumn{2}{|l|}{\textbf{Groundhog RNN}} & \multicolumn{3}{|c|}{\textbf{Best RNN}} & \multicolumn{2}{|l|}{\textbf{Transformer}} \\ \hline
        ~ & \textbf{en-de} & \textbf{Iv-en} & \textbf{layers} & \textbf{en-de} & \textbf{Iv-en} & \textbf{en-de} & \textbf{Iv-en} \\ \hline
        \hline
        NeuralMonkey & 13.7 & 10.5 & 1/1 & 13.7 & 10.5 & - & - \\ 
        OpenNMT-Py & 18.7 & 10.0 & 4/4 & 22.0 & 13.6 & - & - \\ 
        Nematus & 23.9 & 14.3 & 8/8 & 23.8 & 14.7 & - & - \\ 
        Sockeye & 23.2 & 14.4 & 4/4 & 25.6 & 15.9 & 27.5 & 18.1 \\ 
        Marian & 23.5 & 14.4 & 4/4 & 25.9 & 16.2 & 27.4 & 17.6 \\ 
        Tensor2Tensor & - & - & - & - & - & 26.3 & 17.7 \\ \hline
        \textbf{Joey NMT} & 23.5 & 14.6 & 4/4 & 26.0 & 15.8 & 27.4 & 18.0 \\ \hline
    \end{tabular}
    \caption{Results on WMT17 newstest2017}\label{tbl:WMT17}
\end{table*}
In machine translation, active learning was first applied to SMT \citep{haffari-etal-2009-active}, which proposed statistical algorithms and the effectiveness of active learning from the perspective of data coverage. With a large monolingual corpus availability, an active learning strategy is employed to select the most informative sentences for human translation.

The main aim of the current work is to extend Joey NMT, a neural machine translation toolkit based on PyTorch, with active learning techniques for low-resource language translation using the English-Hindi language corpus. We implemented two active learning sampling strategies, least confidence and margin sampling, to obtain the most useful samples to be supervised. First, we trained the transformer-based NMT architecture to obtain the baseline model. We added the active learning technique to generate active learning NMT models. Further, we trained our transformer-based NMT model with the data we provided to our baseline and active learning models to evaluate the performance of the active learning models. As a result of this, we have four models, baseline model, fully trained model, active learning least confidence based model, and active learning margin sampling based model. We also used Byte Pair Encoding (BPE) to enable open vocabulary translation and analysis on the grounds of different BLEU scores to improve the quality of the existing NMT output \citep{Brown1993} \citep{ramanathan-etal-2011-clause}. We discuss the literature survey in the following section and briefly describe our baseline and active learning model architecture methodology. We further explain the experimental settings used in this research. Then, we document a comparative analysis of results acquired from various models. Finally, we end the document with a conclusion and remarks on future scope of active learning in machine translation.

% === II. Literature Survey ========================
% =================================================================================
\section{Literature Survey}
\subsection {Machine Translation}

% % =======
% % FIG. 01
% % =======
% \begin{figure}
%   \begin{center}
%   \includegraphics[width=3.5in]{pdf/01.pdf}\\
%   \caption{Microwave rectifier circuit diagram. An ideal blocking capacitor $C_b$ provides DC isolation between the microwave source and rectifying element.  An ideal choke inductor $L_c$ isolates the DC load $R_{DC}$ from RF power.}\label{circuit_diagram}
%   \end{center}
% \end{figure}

Machine Translation is a branch of computational linguistics that uses a computing device to convert text between languages. Machine translation was first presented by Petr Petrovich Troyanskii \citep{Hutchins-2000}. Machine translation has been thoroughly researched using various models. Rule-based systems were the subject of earlier studies, which gave space to example-based systems in the 1980s. Starting in the late 1980s, we could see that statistical machine translation attained popularity, and various word-based and phrase-based methods requiring little or no linguistic knowledge were implemented. The application of deep neural networks in machine translation became a significant field of research after introducing deep neural networks in 2012.

\subsection{Neural Machine Translation}

Neural machine translation (NMT) is a machine translation technique that uses an artificial neural network to estimate the likelihood of a word sequence and often models full sentences in a single integrated model. A neural network-based translation approach has been used to overcome Statistical Machine Translation (SMT) limitations, such as accuracy and context analysing ability \citep{inproceedings}. NMT use only a fraction of the memory compared to conventional SMT models. Additionally, unlike conventional translation systems, the neural translation model is trained end-to-end to maximise translation performance \citep{kalchbrenner-blunsom-2013-recurrent} \citep{arxiv.1409.3215} \citep{arxiv.1409.1259}. NMT is based on a simple encoder-decoder based network. The encoder aims to decompose the sentence logic and word pairs into embeddings, which can then be stored. The decoder then uses those embeddings to produce the translated sentence. NMT models require a wide corpus of training data based on translations or annotated data created by language specialists. The data used for training in popular languages like English, Spanish, and French has already been processed in huge amounts. Nevertheless, little or no translated data is available for less popular languages and dialects. Unique architectures and methods are required to support low-resource language NMT. The vanilla architecture of encoder-decoder mechanism \citep{arxiv.1409.3215} is commonly used in modern NMT models. The encoder-decoder mechanism learns to optimise conditional log-likelihood after being jointly trained. Numerous encoder-decoder architectures have been created, each modelling the probability distribution differently. In this research, we are using Joey NMT which is built on encoder-decoder architecture.

\subsubsection{Joey NMT}

Joey NMT \citep{hieber-etal-2018-sockeye} is a PyTorch-based, simple neural machine translation toolkit. Joey NMT provides many popular NMT features in a small and simple code base. Despite its focus on simplicity, Joey NMT supports standard network architectures (RNN, transformer, different attention mechanisms, input feeding, configurable encoder/decoder bridge), label smoothing, standard learning techniques (dropout, learning rate scheduling, weight tying, early stopping criteria), beam search decoding, an interactive translation mode, visualization/monitoring of learning progress and attention, checkpoint averaging, and more, and achieves performance comparable to more complex toolkits on standard benchmarks. Table \ref{tbl:WMT17} shows that Joey NMT performs very well compared to other shallow, deep and transformer models. This experiment was conducted using the settings of \citep{hieber-etal-2018-sockeye}, using the exact same data, pre-processing, and evaluation using WMT17-compatible Sacre- BLEU scores \citep{arxiv.1804.08771}.
\begin{table}[h!]
\begin{center}
\begin{tabular}{|l|l|}
\hline
\textbf{System} & \textbf{de-en}\\
\hline
\hline
\citep{wiseman-rush-2016-sequence} & 22.5\\
\citep{bahdanau2016actor} & 27.6\\
\textbf{Joey NMT} (RNN, word)  & 27.1\\
\textbf{Joey NMT} (RNN, BPE32k)  & 27.3\\
\textbf{Joey NMT} (Transformer, BPE32k)  & 31.0\\ \hline
\end{tabular}
\caption{Table  2 IWSLT14 test results.}\label{tbl:IWSLT14}
\end{center}
\end{table}

In another study, data, pre-processing, and word-based vocabulary of Wiseman and Rush \citep{wiseman-rush-2016-sequence} and evaluated with SacreBLEU \citep{arxiv.1804.08771}. Table \ref{tbl:IWSLT14} shows that Joey NMT performs well here, with both recurrent and  Transformer models.

JoeyNMT toolkit has been used to implement the transformer and RNN \citep{arxiv.1907.12484}.

\subsection{Active Learning}

Active learning is a semi-supervised machine learning to select the most informative sentences that need to be labelled to have the highest impact on training a supervised model using acquisition functions. Model-driven and data-driven are the two categories of acquisition functions. All of the techniques we employ for the model-related function are predicated on the concept of uncertainty. We develop a word frequency-based technique that considers linguistic factors for the data driven function. It has been found that active NMT training is advantageous for both varieties of acquisition functions \citep{zhao-etal-2020-active}. In this way, we can build a non-redundant translated corpus on which NMT can be trained to achieve a better performance than models trained on randomly built corpora. The selective sampling approach proposed by \citep{Cohn94improvinggeneralization} is based on the principle of membership questions. However, from an example-driven perspective, the learner asks the instructor about data that it is unsure about, data about which it thinks misclassifications are feasible \citep{Olsson2009}. In the next section, various approaches of active learning have been discussed.

\subsubsection{Query by uncertainty}

Query by uncertainty such as uncertainty sampling and reduction queries the learning instances about which the present theory is least optimistic. Based on the ideas presented by \citep{Cohn94improvinggeneralization}  the learner chooses which occurrences to question the oracle. First, a single classifier is learned from classified data and then used to examine the unlabeled data in question by uncertainty. Next, a human annotator can classify the instances in the unlabeled data set for which the classifier is least certain. The third stage involves the use of trust ratings. The base learner must have a score showing how sure it is in each prediction it makes in this simple process \citep{Olsson2009}.

\subsubsection{Query by committee}

Query by committee is a selective sampling procedure similar to query by uncertainty, with the only difference being that query by committee is a multi-classifier technique. First, several queries are randomly sampled from the version space in the initial conception of the query by committee \citep{sos-qc-92}. The committee is then used to review the collection of unlabeled data. Finally, the difference between the hypotheses about the class of a given instance is used to determine whether or not the human annotator can classify that instance. In the original context, query by committee is only available with base learners for whom access and sampling from the version space are possible.\citep{Liere1997ActiveLW} \citep{Freund2004SelectiveSU} \citep{Olsson2009}.

\subsubsection{Active learning with redundant views}

Using redundant views is similar to the query by committee approach specified above. However, redundant views mean dividing the feature set into many sub-sets or views. Each is sufficient to explain the underlying problem to some degree, rather than arbitrarily sampling the version space or otherwise tampering with the original training data to expand it to achieve a committee \citep{Olsson2009}.

\subsubsection{Related work}

In terms of NLP, named entity recognition and text classification using active learning have been extensively researched \citep{arxiv.1707.05928}. \citep{peris-casacuberta-2018-active} used acquisition functions based on attention for NMT. Reinforcement learning was introduced to actively train an NMT model by \citep{Liu2018}. One study proposes two new and effective sentence selection techniques for active learning: selection based on semantic similarity and decoder probability. Experiments on Indonesian-English and Chinese-English show that selection approaches are superior to random selection and two conventional selection methods \citep{8629116}. Further, work has been conducted in which a comprehensive evaluation of different active learning algorithms on a publicly available dataset (WMT’13) using the SotA NMT architecture has been done \citep{zeng-etal-2019-empirical}. Information retrieval, named object identification, document categorization, part-of-speech marking, decoding, word meaning disambiguation, spoken language comprehension, phone sequence recognition, automated transliteration, and sequence segmentation have all been effectively used with active learning \citep{Olsson2009}.

% === III. Methodology=======================================
% =================================================================================
\section{Methodology}

In this section, we outline the NMT and Active Learning NMT architecture. First, we discuss the RNN and transformer-based NMT models and how we use them in baseline models. Then, we describe how we incorporate an active learning framework in our baseline NMT models. In an active learning framework, we utilise oracle for labelling new data points. We propose acquisition functions, which have been most commonly worked upon in active learning, where the learner queries the instance about which it has the least certainty \citep{Scheffer2001ActiveHM} \citep{culotta2005reducing} \citep{kim-etal-2006-mmr}. These queries technique are based on a model-driven approach. The model, the labelled dataset, and the unlabeled dataset are all used in model-driven techniques to sample sentences.
These techniques receive direct input from the model, which may aid in sampling more sentences from weakly modelled areas of the input space. We describe two model-driven approaches, least confidence and margin query, which select sample instances where the model is least certain about the prediction. Further, to understand how our baseline and active learning models perform, we utilise evaluation metrics like BLEU (Bilingual Evaluation Understudy) and perplexity score described in \citep{papineni2002bleu} \citep{jelinek1977perplexity} are governing metrics for NMT tasks.

\subsection {NMT Architecture}
The purpose of NMT, a specific form of sequence-to-sequence learning, is to produce another sequence of words in the target language from a source language word sequence. This work uses autoregressive recurrent and fully attentional models from Joey NMT. In this, a source sentence of length \(lx\) is represented by a sequence of one-hot encoded vectors \( x_1, x_2, . . , x_{l_x} \) for each word. Analogously, a target sequence of length \(l_y\) is represented by a sequence of one-hot encoded vectors \( y_1, y_2, . . , y_{l_y} \).

\subsubsection{RNN}

The baseline NMT architecture implements RNN encoder-decoder variant from \citep{luong-etal-2015-effective}. The embeddings matrix \(E_{src}\) and a recurrent computation of states allow the encoder RNN convert the input sequence \( x_1, x_2, . . , x_{l_x} \) into a sequence of vectors \( h_1, h_2, . . , h_{l_x} \). 

\[ h_i = RNN(E_{src}x_i , h_{i-1})\]         
\[h_0 = 0\]

We could either use LSTM or GRU to built RNN model. Hidden states from both sides are combined to generate \(h_i\) for a bidirectional RNN. A vector of zeros makes up the initial encoder hidden state, or \(h_0\). Each resulting output sequence, \( h_1, h_2, . . , h_{l_x} \), can be used as the input to the subsequent RNN layer to create several layers.  The decoder employs input feeding, in which an attentional vector \(\tilde{s}\) is joined to the representation of the preceding word as input to the RNN. Decoder states are calculated as follows:
\begin{equation}
\mathbf{s}_t=\operatorname{RNN}\left(\left[E_{t r g} \mathbf{y}_{t-1} ; \tilde{\mathbf{s}}_{t-1}\right], \mathbf{s}_{t-1}\right)\nonumber
\end{equation}

\begin{equation}
\mathbf{s}_0= \begin{cases}\tanh \left(W_{\text {bridge }} \mathbf{h}_{l_x}+\mathbf{b}_{\text {bridge }}\right) & \text { if bridge } \\ \mathbf{h}_{l_x} & \text { if last } \\ \mathbf{0} & \text { otherwise }\end{cases}\nonumber
\end{equation}

\begin{equation}
\tilde{\mathbf{s}}_t=\tanh \left(W_{a t t}\left[\mathbf{s}_t ; \mathbf{c}_t\right]+\mathbf{b}_{a t t}\right)\nonumber
\end{equation}
The starting decoder state can be set to be a vector of zeros, a non-linear transformation of the last encoder state (referred to as "bridge"), or the same as the last encoder state (referred to as "last"). The previous decoder state \(s_{t-1}\) and each encoder state \(h_i\) are scored by an attention mechanism, and the scoring function is either a multi-layer perceptron \citep{arxiv.1409.0473} or a bilinear transformation \citep{luong-etal-2015-effective}. A vector \(o_t = W_{out}\)\(\tilde{s_t}\) , which holds a score for each token in the target language, is created by the output layer. These scores can be understood as a probability distribution over the target vocabulary $\mathcal{V}$ that defines an index over the target tokens \(v_j\) using a softmax transformation \citep{arxiv.1907.12484}. 

\begin{equation}
p\left(y_t=v_j \mid x, y_{<t}\right)=\frac{\exp \left(\mathbf{o}_t[j]\right)}{\sum_{k=1}^{|\mathcal{V}|} \exp \left(\mathbf{o}_t[k]\right)}\nonumber
\end{equation}
\subsubsection{Transformer}

Joey NMT uses code from The Annotated Transformer \citep{rush-2018-annotated} to implement the Transformer from \citep{arxiv.1706.03762}. First, given the \( x_1, x_2, . . , x_{l_x} \) input sequence, create the matrix $X\in R^{l_x\times d}$, where \(l_x\) is the length of the sentence and \(d\) is the dimensionality of the embeddings. Next, we use \(E_{src}x_i\) to look up the word embedding for each input word, then we apply a position encoding and stack the word embeddings that result. The following learnable parameters are defined:

\[A\in R^{d\times d_a}\quad   B\in R^{d\times d_a}\quad   C\in R^{d\times d_o}\]

where \(d_o\) is the output dimensionality, and \(d_a\) is the attention space's dimension. These matrices transform the input matrix into new word representations (\(H\)) by paying attention to all other source words.

\[H = softmax(XAB^TX^T)XC\]

Multi-headed attention is implemented using NMT, where this transformation is computed \(k\) times, once for each head, with various parameters \(A\), \(B,\) and \(C\). We concatenate the results of computing all \(k\) \(H\)s in parallel, apply layer normalisation, and then add a final feed-forward layer.

$$H = [H^{(1)};...;H^{(k)}]$$
\[H^{'} = layer\mbox{-}norm(H) + X\]
\[H^{(enc)} = feed\mbox{-}forward(H^{'}) + H^{'}\]$
$
To ensure that \(H\in R^{l_x\times d}\), we set \(d_o = d/k\). By setting \(X = H^{(enc)}\) and rerunning the calculation, several layers can be piled on top of one another. In contrast to the encoder, the transformer decoder receives as input the stacked target embeddings \(Y\in R^{l_y\times d}\).

\[H = softmax(YAB^TY^T)YC\]
Setting those attention scores to \(-inf\) before the softmax for each target position prevents the user from paying attention to subsequent input words. We compute multi-headed attention again, but this time between intermediate decoder representations \(H^{'}\) and final encoder representations \(H^{(enc)}\), after obtaining \(H^{'} = H + Y\) and before the feed-forward layer.

\[Z = softmax(H^{'}AB^TH^{(enc)T})H^{(enc)}C\]
\[H^{(dec)} = feed\mbox{-}forward(layer\mbox{-}norm(H^{'} + Z))\]

Using \(H^{(dec)}W_{out}\), we predict the target words. 

\subsection{Active Learning NMT}

When dealing with low-resource language, it becomes prohibitively expensive to train the NMT model. Aiming to address this problem, in the AL framework, an acquisition function selects a subset of sentences worth being selected for NMT training. Streaming and pooling are the procedures through which we could send the data to our acquisition function. In a streaming scenario, the acquisition function is presented with training samples, one at a time. The acquisition function will either ignore the sample or send the sample for the query to its label. In a pooling scenario, the acquisition function assesses the log loss probabilities for the unlabeled data and chooses a portion of it for oracle labelling. We select the pooling method as it is more practical to send batches of data for labelling rather than a single sentence at disjoint intervals of time.

\subsubsection{Oracle}

Oracle plays a crucial part in a machine learning task. When given a source sentence for NMT, an oracle can produce the ground truth translation (specifically, an expert human translator). A parallel corpus is gradually constructed using an oracle to translate the selected sentences. In our study, we created the algorithm such that the oracle could be either the human annotator or extract the corresponding target sentences from the parallel corpus if we set the interaction parameter as False. In our study, unlabeled data is the source sentences of a parallel corpus whose target sentences we hide. We extract the corresponding target sentences to label new data points from unlabeled data. We could specify the interaction as True if we want labels from a human annotator.

\subsubsection{Acquisition Function}

Sentences with higher scores are more likely to be chosen as the training corpus. There are two categories of acquisition functions: model-driven and data-driven. A model-driven acquisition function uses a sentence as the model input and output and assigns a score accordingly. The informativeness of the sentence itself is frequently a concern of a data-driven acquisition function, which can score each sentence before training the model. 

\subsubsection{Active Learning Framework}
\label{marker}
Algorithm 1 describes the active learning NMT implementation \citep{zeng-etal-2019-empirical}. It anticipates a labelled parallel corpus ($\mathcal{L}$) for training the baseline NMT system ($\mathcal{M}$), an unlabeled monolingual corpus ($\mathcal{U}$) for sampling new data points for translation and an acquisition function \(\varphi(.)\) for estimating the significance of data points in ($\mathcal{U}$) and batch size ($\mathcal{B}$) for selecting the number of data points to sample in each iteration. The budget is the number of the queries iterations set during training process. Our experiment already has reference translations for all the unlabeled data points. We repeat this process until we have used all the data points in $\mathcal{U}$. We initially train an NMT system with $\mathcal{L}$ for each iteration. Then, using an acquisition function that accounts for $\mathcal{L}$, $\mathcal{U}$, and $\mathcal{M}$, we assign a score to each sentence in $\mathcal{U}$. The next section goes into great detail about the acquisition function and its variations, which is a crucial part of all active learning algorithms. Finally, each sentence in the monolingual source corpus is scored using an acquisition function.

\vspace{5mm} %5mm vertical space

\underline{\textbf{Algorithm 1} Batch Active Learning for NMT}

1: \textbf{Given}: Parallel data $\mathcal{L}$, 

\quad Monolingual source language data $\mathcal{U}$,

\quad Sampling strategy $\psi(\cdot)$, 

\quad Sampling batch size $\mathcal{B}$. 

2: \textbf{while} Budget $\neq$ EMPTY \textbf{do}

3: $\quad \mathcal{M}=$ Train $N M T$ system $(\mathcal{L}) ;$

4: $\quad \textbf{for} \;x \in \mathcal{U} \; \textbf{do} $

5: $\quad \quad f(x)=\psi(x, \mathcal{U}, \mathcal{L}, \mathcal{M})$

6: $\quad \textbf{end \;for} $

7: $\quad X_B=$ TopScoringSamples $(f(x), \mathcal{B}) ;$

8: $\quad Y_B=$ HumanTranslation $\left(X_B\right)$

9: $\quad \mathcal{U} = \mathcal{U} - X_B; $

10: $\quad \mathcal{L} = \mathcal{L} \cup \;\{X_B, Y_B\}; $

11: $\textbf{end while} $

12: $\textbf{return} \;\mathcal{L} $

\vspace{5mm} %5mm vertical space
Then, the best $\mathcal{B}$ sentences are picked to be translated. Finally, the references translations of these sentences are added to $\mathcal{L}$ along with their removal from $\mathcal{U}$. As a result, a parallel corpus is gradually constructed by utilising an oracle to translate the sentences with high scores. After that, the process continues for the specified amount of iterations. The NMT model is then retrained using the parallel corpus. As a result, the model is trained on those labelled examples before being evaluated to see how well it is doing. As a result, new data is gradually added to the NMT system. 

\subsection{Model-driven Query Strategies}

One of the key elements of active learning is to have a meaningful strategy for obtaining the most useful samples to be supervised. For this, we require an evaluation of the informativeness of unlabeled samples. Model-driven approaches estimate the prediction uncertainty of a source sentence given the machine translation model parameters and select sentences with high uncertainty for training the model. The sampling strategies used in this work are based on uncertainty. Settles and Craven \citep{settles-craven-2008-analysis} tried these methods on sequence labelling tasks. The idea behind the uncertainty sampling method is to select those instances for which the model has the least confidence to be correctly translated. Therefore, all techniques compute, for each sample, an uncertainty score. The selected sentences will be those with the highest scores.

\subsubsection{Least Confidence}

\citep{culotta2005reducing} describe a simple uncertainty-based approach for least confidence-based sequence models. The active learner is least certain of its prediction for the most likely label in the instance chosen by the least confident sampling technique. For sequence-based models, uses an earlier least confidence sampling strategy where the input sequence is designated by \(x\) and the label sequence is represented by \(y\). Its query strategy formulation \(\phi^{LC}{(x)}\) can be written as follows: 

\[\phi^{LC}{(x)} = 1 - P(y^{*}|x;\theta).\]

Where most likely label sequence according to the learner is represented by \(y*\) and posterior probability of \(y\) given x is denoted by \(P(y^{*}|x;\theta).\)

\subsubsection{Margin Sampling}

Another uncertainty technique put forth by \citep{Scheffer2001ActiveHM} involves querying the instance with the smallest margin between the posteriors for its two most likely labellings. This strategy is known as margin \((M)\).

\[\phi^{M}{(x)} = -(P(y_1^{*}|x;\theta)-P(y_2^{*}|x;\theta).\]

Here,  the first and second best label sequences are \(y_1^*\) and \(y_2^*\), respectively. Their posterior probabilities are by \(P(y_1^{*}|x;\theta)\) and \((P(y_2^{*}|x;\theta)\) respectively. The model cannot distinguish between the best and inferior translations with a narrow margin. This concept is incorporated into active learning to select the samples.

\subsection{Evaluation}

Evaluation is highly challenging in various NLP tasks and non-trivial. In NMT, we evaluate by comparing the model’s hypothesis with the actual reference sentence. We should only compare scores between a language and not across different languages. We grade our models using the following two metrics.

\subsubsection{Bilingual Evaluation Understudy (BLEU)}

Bilingual Evaluation Understudy (BLEU) is the standard evaluation metric in NMT proposed by \citep{papineni2002bleu}. This technique is less expensive, quicker, and linguistically unrestricted than human evaluation. It is an algorithm that evaluates the quality of the machine-translated text. The main idea behind BLEU is that the closer a machine translation is to a professional human translation, the better it is. BLEU calculates an average mean of the precision of the n-grams from the hypothesis that appear in the reference sentence. BLEU employs a modified form of precision to compare output text to various reference phrases. The reference sentences are human-translated text. Additionally, it imposes a brevity penalty on short translations.
Usually given as a value between 0 and 1, output values can be quickly changed to percentages if necessary. A larger number of reference sentences will result in higher BLEU scores. A higher BLEU score indicates greater machine translation quality. BLEU is computed using a couple of ngram modified precisions. Specifically,

\begin{equation}
BLEU=BP\cdot{\exp{({\sum_{n=1}^{N}w_n\log{p_n}})}}\nonumber
\end{equation}

where $p_n$
is the modified precision for ngram, the base of log is the natural base e, $w_n$ is weight between 0 and 1 for $\log{p_n}$ and $\sum_{n=1}^{N}w_n=1$
, and BP is the brevity penalty to penalize short machine translations.

\begin{equation}
BP= 
\begin{cases}
1 & \text { if c$>$r }\\ 
{\exp(1- r/c)} & \text { c$\le$r }\end{cases}\nonumber
\end{equation}

where c
is the number of unigrams (length) in all the candidate sentences, and r is the best match lengths for each candidate sentence in the corpus. Here the best match length is the closest reference sentence length to the candidate sentences. 

\subsubsection{Perplexity}

How well a probability distribution or probability model predicts a sample is measured by perplexity. Perplexity is a metric used in natural language processing to assess language models. A language model is a probability distribution applied to complete texts or sentences.  A low perplexity value suggests that the probability distribution may accurately predict the sample. The reciprocal of the (geometric) average probability that the model allocated to each word in the test set \(T\) is the perplexity \(PP_p{(T)}\) of the model p. It is related to cross-entropy by the below equation.

\[PP_p{(T)} = 2^{H_{p}(T)}\]
 
where \(H_p\) is cross-entropy. Lower cross-entropies and perplexities are preferable \citep{Brown1992}.

% === IV. Experiment ========================================
% =================================================================================
\section{Experiments}
\subsection{Dataset}
In this research, we use English-Hindi language pair dataset. For training our NMT models, we use  IIT Bombay English-Hindi Parallel Corpus \citep{kunchukuttan-etal-2018-iit}. The parallel corpus has been compiled from a variety of existing sources (primarily OPUS \citep{tiedemann-2012-parallel}, HindEn \citep{bojar-etal-2014-hindencorp} and TED \citep{abdelali-etal-2014-amara}) as well as corpora developed at the Center for Indian Language Technology (CFILT), IIT Bombay over the years. The corpus consists of sentences, phrases, and dictionary entries, spanning many applications and domains. This data is publicly available and we use an open-source platform, Huggingface, as it provides a consistent way of accessing data using an API. This reduces the burden on our end to maintain a data repository. The dataset can also be accessed in raw format from IIT Bombay online repository\footnote{http://www.cfilt.iitb.ac.in/iitb\_parallel}. The training, dev and test corpora consist of 1.6 million, 520 and 2507 sentences of English and Hindi Language pair. We are not using dev and test corpora becuase of its small size. We construct our own dev and test corpora from a 1.6 million dataset in the data preprocessing stage, with 40K sentences in each, see data split in table \ref{Full dataset split}.

\begin{table}[htb]
\bigskip
\begin{center}
\begin{tabular}{|l|l|l|l|}
\hline
\textbf{Training Data} & \textbf{Dev Data} & \textbf{Test Data} & \textbf{Total Data}  \\
\hline
\hline
1552563 & 40856 & 40858 & 1634277 \\
\hline
\end{tabular}
\end{center}
\caption{Full dataset split}
\label{Full dataset split}
\end{table}
For active learning we randomly picked 30\% of data from training set. See table \ref{Active learning dataset split} 
\begin{table}[htb]
\bigskip
\begin{center}
\begin{tabular}{|l|l|}
\hline
\textbf{Data Split} & \textbf{Size}\\\hline
\hline
Baseline Training Data & 1086795\\\hline
Active Learning Training Data & 465768\\\hline
Dev Data & 40856 \\\hline
Test Data & 40858 \\\hline
Total Data & 1634277\\
\hline
\end{tabular}
\end{center}
\caption{Active learning dataset split}
\label{Active learning dataset split}
\end{table}

\subsection{Data Preprocessing}

The first step of after acquiring data from Huggingface Data API is to perform the pre-processing of data in which we prepare and cleans the dataset and reduces the noise of the dataset. This task included the conversion of all sentences into lowercase, removing special and bad characters, removing stop words, removing extra white-spaces and tabs, remove characters not related to language like we came across Urdu characters in our Hindi corpus. Further, to avoid data leakage, we checked missing and duplicated sentences and ensured test data was filtered from the training and dev sets.

We use the IIT Bombay English-Hindi Parallel Corpus consisting of 1.6 M sentence pairs. In order to remove bias, the data was shuffled and  test and validation split of 40k each parallel corpus sentences were used. The remaining parallel corpus was utilised for training the models. From the training data, we randomly sampled 70\% of the whole bilingual training dataset, this we called Baseline Training Data,(see \ref{Active learning dataset split}). We used Baseline Training Data to train an initial NMT model (Baseline NMT model) and the remaining data was purposed as active learning corpus ($\sim$465k) used for simulating the active learning experiments. We performed random sampling initially and fix the labelled and unlabelled datasets for all the experiments for a fair comparison. 

Since, we experiment in a simulated active learning framework, the target sentences in active learning dataset are hidden while scoring source sentences with different active learning strategies. Once the active learning algorithm samples a batch from 465k source sentences from active learning dataset, the sampled sentences, along with their corresponding “hidden” translations, are added to labelled dataset. We calculated the various data statistics to know our sentences better, which includes the vocabulary size and length of the sentences. The vocabulary size for English and Hindi languages is 21K and 37K respectively. The table \ref{tabel:Sentence length} shows the summary data statistics.
\begin{table}[htb]
\bigskip
\begin{center}
\begin{tabular}{|l|l|l|l|}
\hline
 & \textbf{Max length} & \textbf{Avg length} & \textbf{Median length}  \\
\hline
\hline
en & 1681.0 & 13.94 & 10.0 \\
hi & 1291.0 & 13.74 & 10.0 \\
\hline
\end{tabular}
\end{center}
\caption{Sentence length}
\label{tabel:Sentence length}
\end{table}
Next, the key function preprocessing step is to tokenise source and target sentences and create a dictionary, which indexes the words in the training process. The dictionary lists out all unique words. 
We used Moses \citep{koehn2007moses} toolkit for tokenisation and cleaning the English dataset. The Hindi dataset is first normalised with Indic NLP library\footnote{https://anoopkunchukuttan.github.io/indic\_nlp\_library/}, followed by tokenisation with the same library.
Spell normalisation is cardinal while processing the Hindi text. A single word in Hindi can be written in many forms with the same underlying meaning. In normalisation stage, all such similar words are mapped to a single word to mitigate lexical redundancy. We employ byte-pair encoding (BPE) \citep{sennrich-etal-2016-improving} to generate a shared vocabulary for each language pair. It is a BPE tokenisation technique whose definition is based on the number of merges.
The algorithm separates the corpus into words by removing white space, counting all nearby character pairs, combining the characters in the most common pair, and then adding them to the vocabulary. The resulting BPE tokenisation had 16k merge operations.

\subsection{Baseline NMT Experimental Setup}

For all of our experiments, we used JoeyNMT toolkit. We used the Transformer model in our submissions. For our transformer model, we used 6 layers in both encoder and decoder with 256 hidden units in each layer. The word embedding size was set to 256 with 4 heads. We used Byte Pair Encoding (BPE) to learn the vocabulary with 16k merge operations. We used the subword-nmt for learning the BPE vocabulary. Since the writing systems and vocabularies of English and Hindi are separate, BPE models are trained separately. The figure \ref{fig:NMT_1} show how our baseline architecture operates. 
% \begin{figure}
%     \centering
%   \includegraphics[scale=0.1]{figures/nmt_arch.jpg}
%   \caption{Baseline NMT Architecture}
%   \label{fig:NMT_1}
% \end{figure}

The model is trained using Adam optimiser \citep{kingma2014adam} with \(\beta1 =\) 0.9, 
\(\beta2 =\) 0.98 ,
% \(\epsilon = \num{1e-9}\) 
with a learning rate of 0.0003,  and a warm-up for the 1K steps with a maximum length of 60. We used early-stopping based on perplexity(ppl) with patience=5. 
Cross-entropy loss was calculated, and the dropout probability is set to 0.3. We used the minimum learning rate of \(\num{1e-8}\) and the number of epochs to control the training. Here, we also utilised the plateau scheduler, which lowers the initial learning rate by a factor of 0.7 whenever the ppl has not improved for patience validations. Every time a new high score in ppl is attained, checkpoints for the model parameters are saved. We are only keeping the last three best checkpoints to save memory on old checkpoints. Later, we use the best checkpoint parameters to train our active learning NMT model. Xavier weight initialisation has been used. Each layer in our encoder and decoder includes attention sub-layers in addition to a fully linked feed-forward network that is applied to each position individually and identically.
\vspace{2mm} %5mm vertical space
\noindent%
\begin{minipage}{\linewidth}% to keep image and caption on one page
\makebox[\linewidth]{%        to center the image
  \includegraphics[scale=0.4]{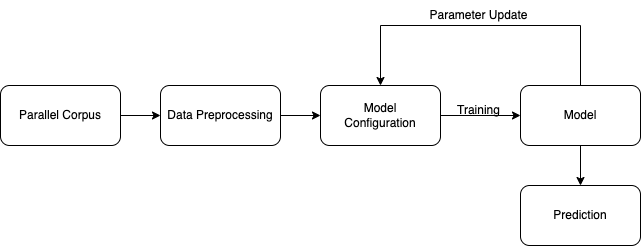}}
\captionof{figure}{Baseline NMT Architecture}\label{fig:NMT_1}%      only if needed  
\end{minipage}
\vspace{2mm} %5mm vertical space

It is composed of two linear transformations separated by a ReLU activation. First, we transform the input and output tokens into vectors of model dimension using learnt embeddings. To translate the output of the decoder into estimated next-token probabilities, we also employ the standard learnt linear transformation and softmax function.
Our model's pre-softmax linear transformation and the two embedding layers share the same weight matrix \citep{arxiv.1608.05859}. We used label smoothing with a value of \(\epsilon(ls)\) = 0.1 during training.
As a result, the model becomes less perplexing, which reduces perplexity but increases accuracy and BLEU score. All models are trained with a batch size of 4096 sentences for 40 epochs. After 1000 mini-batches, the model will get validated using the dev dataset. Using greedy sampling to decode the test dataset, we evaluated our model and ran inference with a beam size of 5 and a batch size of 1024. We use the BELU score and PPL as our evaluation metric.

\subsection{Active Learning NMT Experimental Setup}

The active learning model is a selective sampling technique. Referring to the architecture presented in section, 
\nameref{marker},
in our approach, we first train a baseline model ($\mathcal{M}$). Then, this model is built on top of the NMT model, which utilises base parallel corpus ($\mathcal{L}$) until it acquires a local minimum. Finally, we train this baseline model and set the stage for the active learning model. The figure \ref{fig:NMT_2} show how our Active learning NMT architecture operates. 
% \begin{figure}
%     \centering
%   \includegraphics[scale=0.1]{figures/active nmt.jpg}
%   \caption{Active Learning NMT Architecture}
%   \label{fig:NMT_2}
% \end{figure}

\vspace{2mm} %5mm vertical space
\begin{minipage}{\linewidth}% to keep image and caption on one page
\makebox[\linewidth]{%        to center the image
  \includegraphics[scale=0.4]{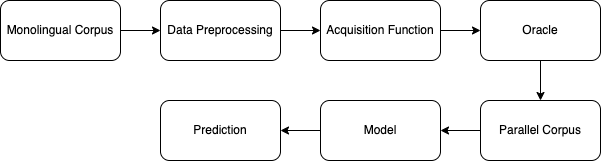}}
\captionof{figure}{Active Learning NMT Architecture}\label{fig:NMT_2}%      only if needed  
\end{minipage}

\vspace{2mm} %5mm vertical space
\begin{figure*}
    \centering
    \includegraphics[width=0.8\textwidth]{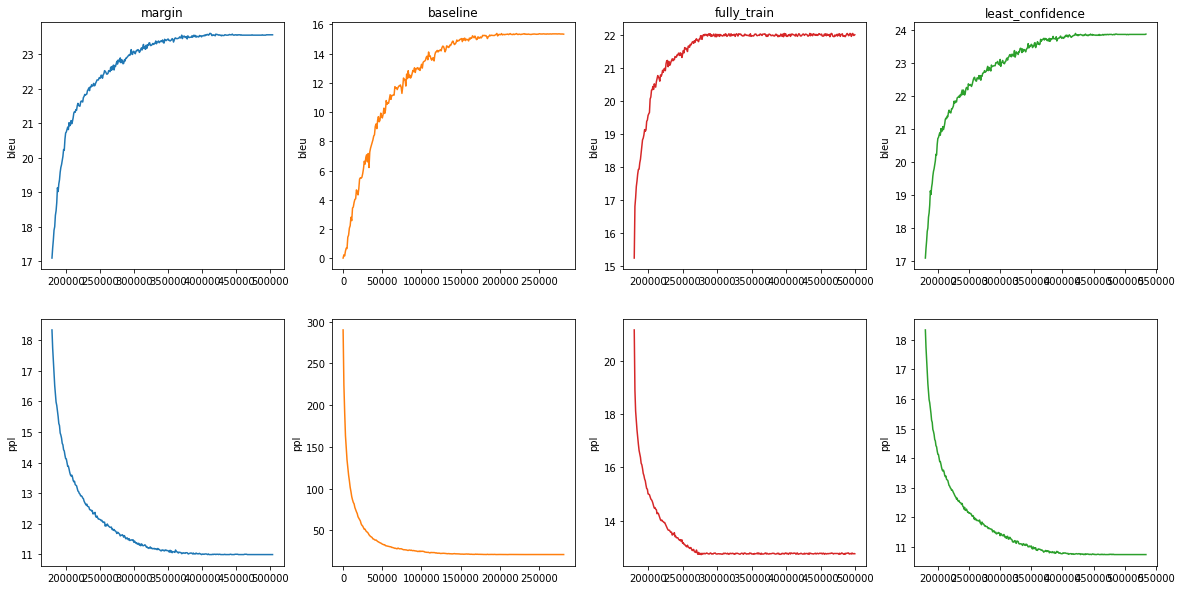}
    \caption{Validation set BELU and Perplexity score during training.}
    \label{fig:Result1}
\end{figure*}
To build the active learning model, we separate the target language and create a monolingual corpus ($\mathcal{U}$) from the active learning corpus. On the subset of this corpus, we do random sampling on 20k samples and find the top N predictions ($X_B$) in the form of log loss probabilities. Then, an oracle iteratively queries these predictions($\mathcal{U}$), applying different acquisition functions. The acquisition function's role is to select the most informative samples based on the algorithm, thresholds and heuristics. The selected samples are then paired up with the actual target translation($\mathcal{L}$). It can be done by oracle either interactively or using the existing parallel corpus. These samples are then fed into the baseline model to improve its score. 

The top N log loss probabilities are selected using the beam search algorithm. The algorithm selects multiple tokens for a position in a given sequence based on conditional probability. It can take any number of N best alternatives through a hyperparameter known as beam width, which in our experiment is set to 5. In addition, we implemented two uncertainty-based acquisition functions, namely least confidence and margin sampling. These sampling techniques are written for a classical problem involving probabilities of different classes. In our work, we transformed and implemented them for machine translations problem.

Active learning is an iterative process, where in each iteration few samples are added to training data. First, the oracle randomly picks up 6\% of the active learning monolingual corpus (pool size), which turns around to be 20K samples. From there, the acquisition function returns 10K samples (query size) which are added to training data each time when oracle is queried. This active learning loop continues for 2 epochs over 20 iterations to the oracle. In each epoch, the training data is added with some part of the active learning data pool until the pool is exhausted. 

In order to add active learning to joeyNMT, we extended JoeyNMT features. JoeyNMT inherits dataset classes from the transformer, we implemented a new class with which we could separate the parallel corpus from the monolingual corpus. The base JoeyNMT architecture had issues with beam search, where it would fail to provide N top log loss probabilities for sentences. It would essentially be fatal for our acquisition; we successfully fixed this obstacle. In addition, the predictions returned by the baseline architecture did not support batch loss reporting. We added this functionality to allow the acquisition function to perform and scale in order of hundreds. Like the baseline model provides different execution modes such as train, test and translate, we created another mode called active learning mode, which is an addition to the existing pipeline and extends the functionality of the baseline joeyNMT. We use BLEU scores and perplexity(ppl), on the same dev and test dataset for all our models to perform validation and evaluation, which we used in our baseline model. 

\section{Result and Analysis}
We evaluate the effectiveness of active learning on two parameters: bleu and perplexity score. In NMT tasks, these are standard metrics, unlike other classical neural networks, which are based on accuracy, precision and recall performance parameters. The figure \ref{fig:Result1} shows model-wise bleu and perplexity scores. All the models are trained on 20 epochs. BLEU score for the fully trained model peaks at 22.56, while the baseline model, which trains on 70\% of the trainset, struggles and attains a local minima of around 16.26.

% \begin{figure}
%     \centering
%   \includegraphics[scale=0.35]{figures/fig9.png}
%   \caption{Validation set BELU and Perplexity score during training.}
%   \label{fig:Result1}
% \end{figure}

Active Learning models, margin and least confidence are built on top of the baseline model. It is clearly depicted in the figure \ref{fig:Result1} and the table \ref{Validation}, they are more performant. The BLEU scores of both models, signal that providing sentences that perform poorly on the model gives us better performances and help model attain local minima quickly.

The table \ref{Validation} describes our final findings on the test dataset, and it gives us a good contrast of how active learning has achieved better results without training on a full data set. We only use $\sim$40,000 samples of  4,65,768 total active learning data samples and have achieved a better performing model at a lesser cost both in training time and in less amount of data. 
\begin{figure*}
    \centering
    \includegraphics[width=0.9\textwidth]{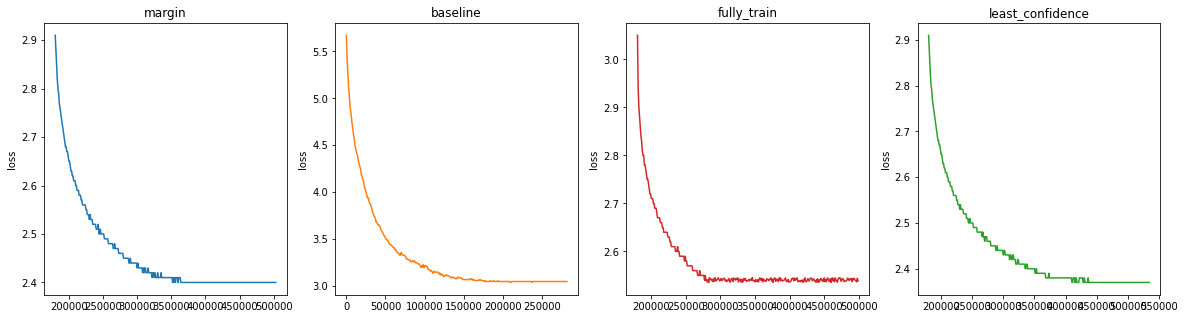}
  \caption{Training Loss comparison.}
  \label{fig:Training Loss comparison}
\end{figure*}
\begin{table}[htb]
\bigskip
\begin{center}
\begin{tabular}{|l|l|l|l||l|}
\hline
 & \textbf{Fully Trained} & \textbf{Baseline} & \textbf{Margin} & \textbf{Least Confidence}\\
\hline
\hline
BLEU & 22.56 & 16.26 & 24.20 & 24.54 \\
Perplexity & 12.71 & 21.00 & 11.71 & 11.70 \\
\hline
\end{tabular}
\end{center}
\caption{Test set BLEU and perplexity score after training.}
\label{Validation}
\end{table}

\begin{figure}[H]
    \centering
  \includegraphics[scale=0.6]{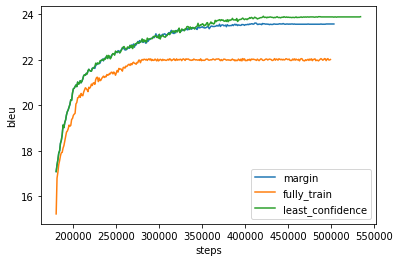}
  \caption{Model comparison.}
  \label{fig:Model comparison on BLEU}
\end{figure}

The figure \ref{fig:Model comparison on BLEU} gives us a better understanding of how different models perform against each other on different steps of epochs on the validation dataset. It is evident that margin sampling and least confidence attain better BELU scores early than their fully trained model counterpart. Furthermore, the least confidence model has a better learning capability than the margin sampling model.  
\begin{figure}[H]
    \centering
  \includegraphics[scale=0.6]{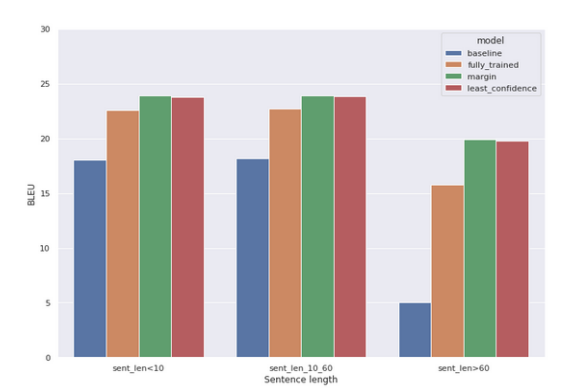}
  \caption{BLEU score for different sentence length.}
  \label{fig:BLEU}
\end{figure}

To understand the learnability of models, we deep dive into the training loss and pick out peculiar patterns. In figure \ref{fig:Training Loss comparison}, the baseline model loss describes an exponential decay as it becomes almost constant around 150K steps, describing that the model stopped learning around 200K. This pattern is acceptable as we had only provided 70\% of training data. A fully trained model and other active learning models appear to learn even 500K steps which means that model convergence is good and a good scope of learning is achieved. 

Next, to understand our models' capabilities and performance on different sentence lengths, we run our models over different test datasets. In figure \ref{fig:BLEU}, we carefully analyze that the active learning models (least confidence and margin) fare better than the fully trained and baseline model. The models are trained on a sentence length of 60 tokens. This provide us concrete evidence that  the baseline and others models have low performance. Margin and least confidence models give similar BLEU scores, but, on deeper inspection, the margin sampling outranks the least confidence.

\section{Conclusions}

In this work, we demonstrate an active learning strategy for incorporating knowledge from monolingual data into the NMT system using the acquisition function. The idea was to supervise the most useful samples from potentially monolingual data. We developed two uncertainty-based acquisition function algorithms, least confidence and margin sampling and studied their effect on improving the translation quality for low resource language Hindi. The research builds upon previous research in the field, using the transformer architecture \citep{arxiv.1706.03762} \citep{rush-2018-annotated} and data to translate English to Hindi. Our experiment results strongly prove that active learning benefits NMT for low-resource languages. Further, the results show improvements in the previous BLEU scores obtained for our parallel corpus by a large margin \citep{kunchukuttan-etal-2018-iit}. Moreover, we obtained consistent reductions of approximately a 25\% of the effort required for reaching the desired translation quality.

\subsection{Future Work}

In the future, we plan to investigate various areas of work.
To see if the findings of this work are still valid, we first hope to apply our methodology to other datasets involving linguistically mixed language pairs. We also wish to research how bandit learning or reinforcement can be included in our architecture. Recent studies \citep{nguyen-etal-2017-reinforcement} that are orthogonal to our work have already demonstrated the value of these learning paradigms. Future work would also involve fine-tuning the training of long and rare sentences using smaller data sets. Finally, since the grammatical structures of many Indian languages are similar, we would like to investigate Active Learning NMT for more low-resource Indian languages like Bengali or Marathi in the future. The code we implemented to train and evaluate our models is available on Github \footnote{https://github.com/kritisingh24/active\_learning\_nmt}.

% or


% ============================================
%\appendices
%Appendix one text goes here %\citep{Roberg2010}.

%\section{}

%\section*{Acknowledgment}

%The authors would like to thank D. Root for the loan of the SWAP. The SWAP that can ONLY be usefull in Boulder...

\ifCLASSOPTIONcaptionsoff
  \newpage
\fi



%\begin{thebibliography}{1}


% \bibliographystyle{plainnat}
\bibliographystyle{cas-model2-names}
\bibliography{IEEEabrv,Bibliography}

\begin{thebibliography}{55}
\expandafter\ifx\csname natexlab\endcsname\relax\def\natexlab#1{#1}\fi
\providecommand{\url}[1]{\texttt{#1}}
\providecommand{\href}[2]{#2}
\providecommand{\path}[1]{#1}
\providecommand{\DOIprefix}{doi:}
\providecommand{\ArXivprefix}{arXiv:}
\providecommand{\URLprefix}{URL: }
\providecommand{\Pubmedprefix}{pmid:}
\providecommand{\doi}[1]{\href{http://dx.doi.org/#1}{\path{#1}}}
\providecommand{\Pubmed}[1]{\href{pmid:#1}{\path{#1}}}
\providecommand{\bibinfo}[2]{#2}
\ifx\xfnm\relax \def\xfnm[#1]{\unskip,\space#1}\fi
%Type = Misc
\bibitem[{IEE()}]{IEEEexample:BSTcontrol}
, .
%Type = Inproceedings
\bibitem[{Abdelali et~al.(2014)Abdelali, Guzman, Sajjad and
  Vogel}]{abdelali-etal-2014-amara}
\bibinfo{author}{Abdelali, A.}, \bibinfo{author}{Guzman, F.},
  \bibinfo{author}{Sajjad, H.}, \bibinfo{author}{Vogel, S.},
  \bibinfo{year}{2014}.
\newblock \bibinfo{title}{The {AMARA} corpus: Building parallel language
  resources for the educational domain}, in: \bibinfo{booktitle}{Proceedings of
  the Ninth International Conference on Language Resources and Evaluation
  ({LREC}'14)}, \bibinfo{publisher}{European Language Resources Association
  (ELRA)}, \bibinfo{address}{Reykjavik, Iceland}. pp.
  \bibinfo{pages}{1856--1862}.
\newblock \URLprefix
  \url{http://www.lrec-conf.org/proceedings/lrec2014/pdf/877_Paper.pdf}.
%Type = Inproceedings
\bibitem[{Ambati et~al.(2011)Ambati, Vogel and Carbonell}]{ambati2011multi}
\bibinfo{author}{Ambati, V.}, \bibinfo{author}{Vogel, S.},
  \bibinfo{author}{Carbonell, J.G.}, \bibinfo{year}{2011}.
\newblock \bibinfo{title}{Multi-strategy approaches to active learning for
  statistical machine translation}, in: \bibinfo{booktitle}{Proceedings of
  Machine Translation Summit XIII: Papers}.
%Type = Article
\bibitem[{Bahdanau et~al.(2016)Bahdanau, Brakel, Xu, Goyal, Lowe, Pineau,
  Courville and Bengio}]{bahdanau2016actor}
\bibinfo{author}{Bahdanau, D.}, \bibinfo{author}{Brakel, P.},
  \bibinfo{author}{Xu, K.}, \bibinfo{author}{Goyal, A.}, \bibinfo{author}{Lowe,
  R.}, \bibinfo{author}{Pineau, J.}, \bibinfo{author}{Courville, A.},
  \bibinfo{author}{Bengio, Y.}, \bibinfo{year}{2016}.
\newblock \bibinfo{title}{An actor-critic algorithm for sequence prediction}.
\newblock \bibinfo{journal}{arXiv preprint arXiv:1607.07086} .
%Type = Misc
\bibitem[{Bahdanau et~al.(2014)Bahdanau, Cho and Bengio}]{arxiv.1409.0473}
\bibinfo{author}{Bahdanau, D.}, \bibinfo{author}{Cho, K.},
  \bibinfo{author}{Bengio, Y.}, \bibinfo{year}{2014}.
\newblock \bibinfo{title}{Neural machine translation by jointly learning to
  align and translate}.
\newblock \URLprefix \url{https://arxiv.org/abs/1409.0473},
  \DOIprefix\doi{10.48550/ARXIV.1409.0473}.
%Type = Article
\bibitem[{Balcan et~al.(2009)Balcan, Beygelzimer and Langford}]{BALCAN200978}
\bibinfo{author}{Balcan, M.F.}, \bibinfo{author}{Beygelzimer, A.},
  \bibinfo{author}{Langford, J.}, \bibinfo{year}{2009}.
\newblock \bibinfo{title}{Agnostic active learning}.
\newblock \bibinfo{journal}{Journal of Computer and System Sciences}
  \bibinfo{volume}{75}, \bibinfo{pages}{78--89}.
\newblock \URLprefix
  \url{https://www.sciencedirect.com/science/article/pii/S0022000008000652},
  \DOIprefix\doi{https://doi.org/10.1016/j.jcss.2008.07.003}.
  \bibinfo{note}{learning Theory 2006}.
%Type = Inproceedings
\bibitem[{Bojar et~al.(2014)Bojar, Diatka, Rychl{\'y}, Stra{\v{n}}{\'a}k,
  Suchomel, Tamchyna and Zeman}]{bojar-etal-2014-hindencorp}
\bibinfo{author}{Bojar, O.}, \bibinfo{author}{Diatka, V.},
  \bibinfo{author}{Rychl{\'y}, P.}, \bibinfo{author}{Stra{\v{n}}{\'a}k, P.},
  \bibinfo{author}{Suchomel, V.}, \bibinfo{author}{Tamchyna, A.},
  \bibinfo{author}{Zeman, D.}, \bibinfo{year}{2014}.
\newblock \bibinfo{title}{{H}ind{E}n{C}orp - {H}indi-{E}nglish and {H}indi-only
  corpus for machine translation}, in: \bibinfo{booktitle}{Proceedings of the
  Ninth International Conference on Language Resources and Evaluation
  ({LREC}'14)}, \bibinfo{publisher}{European Language Resources Association
  (ELRA)}, \bibinfo{address}{Reykjavik, Iceland}. pp.
  \bibinfo{pages}{3550--3555}.
\newblock \URLprefix
  \url{http://www.lrec-conf.org/proceedings/lrec2014/pdf/835_Paper.pdf}.
%Type = Article
\bibitem[{Brown et~al.(1992)Brown, Della~Pietra, Pietra, Lai and
  Mercer}]{Brown1992}
\bibinfo{author}{Brown, P.}, \bibinfo{author}{Della~Pietra, S.},
  \bibinfo{author}{Pietra, V.}, \bibinfo{author}{Lai, J.},
  \bibinfo{author}{Mercer, R.}, \bibinfo{year}{1992}.
\newblock \bibinfo{title}{An estimate of an upper bound for the entropy of
  english}.
\newblock \bibinfo{journal}{Computational Linguistics} \bibinfo{volume}{18},
  \bibinfo{pages}{31--40}.
%Type = Article
\bibitem[{Brown et~al.(1993)Brown, Della~Pietra, Pietra and Mercer}]{Brown1993}
\bibinfo{author}{Brown, P.}, \bibinfo{author}{Della~Pietra, S.},
  \bibinfo{author}{Pietra, V.}, \bibinfo{author}{Mercer, R.},
  \bibinfo{year}{1993}.
\newblock \bibinfo{title}{The mathematics of statistical machine translation:
  Parameter estimation}.
\newblock \bibinfo{journal}{Computational Linguistics} \bibinfo{volume}{19},
  \bibinfo{pages}{263--311}.
%Type = Misc
\bibitem[{Cho et~al.(2014)Cho, van Merrienboer, Bahdanau and
  Bengio}]{arxiv.1409.1259}
\bibinfo{author}{Cho, K.}, \bibinfo{author}{van Merrienboer, B.},
  \bibinfo{author}{Bahdanau, D.}, \bibinfo{author}{Bengio, Y.},
  \bibinfo{year}{2014}.
\newblock \bibinfo{title}{On the properties of neural machine translation:
  Encoder-decoder approaches}.
\newblock \URLprefix \url{https://arxiv.org/abs/1409.1259},
  \DOIprefix\doi{10.48550/ARXIV.1409.1259}.
%Type = Article
\bibitem[{Cohn et~al.(1994a)Cohn, Ghahramani and Jordan}]{cohn1994active}
\bibinfo{author}{Cohn, D.}, \bibinfo{author}{Ghahramani, Z.},
  \bibinfo{author}{Jordan, M.}, \bibinfo{year}{1994}a.
\newblock \bibinfo{title}{Active learning with statistical models}.
\newblock \bibinfo{journal}{Advances in neural information processing systems}
  \bibinfo{volume}{7}.
%Type = Inproceedings
\bibitem[{Cohn et~al.(1994b)Cohn, Ladner and
  Waibel}]{Cohn94improvinggeneralization}
\bibinfo{author}{Cohn, D.}, \bibinfo{author}{Ladner, R.},
  \bibinfo{author}{Waibel, A.}, \bibinfo{year}{1994}b.
\newblock \bibinfo{title}{Improving generalization with active learning}, in:
  \bibinfo{booktitle}{Machine Learning}, pp. \bibinfo{pages}{201--221}.
%Type = Inproceedings
\bibitem[{Culotta and McCallum(2005)}]{culotta2005reducing}
\bibinfo{author}{Culotta, A.}, \bibinfo{author}{McCallum, A.},
  \bibinfo{year}{2005}.
\newblock \bibinfo{title}{Reducing labeling effort for structured prediction
  tasks}, in: \bibinfo{booktitle}{AAAI}, pp. \bibinfo{pages}{746--751}.
%Type = Incollection
\bibitem[{Dagan and Engelson(1995)}]{DAGAN1995150}
\bibinfo{author}{Dagan, I.}, \bibinfo{author}{Engelson, S.P.},
  \bibinfo{year}{1995}.
\newblock \bibinfo{title}{Committee-based sampling for training probabilistic
  classifiers}, in: \bibinfo{editor}{Prieditis, A.}, \bibinfo{editor}{Russell,
  S.} (Eds.), \bibinfo{booktitle}{Machine Learning Proceedings 1995}.
  \bibinfo{publisher}{Morgan Kaufmann}, \bibinfo{address}{San Francisco (CA)},
  pp. \bibinfo{pages}{150--157}.
\newblock \URLprefix
  \url{https://www.sciencedirect.com/science/article/pii/B978155860377650027X},
  \DOIprefix\doi{https://doi.org/10.1016/B978-1-55860-377-6.50027-X}.
%Type = Phdthesis
\bibitem[{Eck(2008)}]{Eck2008_1000010682}
\bibinfo{author}{Eck, M.}, \bibinfo{year}{2008}.
\newblock \bibinfo{title}{Developing Deployable Spoken Language Translation
  Systems given Limited Resources}.
\newblock Ph.D. thesis.
\newblock \DOIprefix\doi{10.5445/IR/1000010682}.
%Type = Article
\bibitem[{Freund et~al.(2004)Freund, Seung, Shamir and
  Tishby}]{Freund2004SelectiveSU}
\bibinfo{author}{Freund, Y.}, \bibinfo{author}{Seung, H.S.},
  \bibinfo{author}{Shamir, E.}, \bibinfo{author}{Tishby, N.},
  \bibinfo{year}{2004}.
\newblock \bibinfo{title}{Selective sampling using the query by committee
  algorithm}.
\newblock \bibinfo{journal}{Machine Learning} \bibinfo{volume}{28},
  \bibinfo{pages}{133--168}.
%Type = Inproceedings
\bibitem[{Guo and Greiner(2007)}]{guo2007optimistic}
\bibinfo{author}{Guo, Y.}, \bibinfo{author}{Greiner, R.}, \bibinfo{year}{2007}.
\newblock \bibinfo{title}{Optimistic active-learning using mutual
  information.}, in: \bibinfo{booktitle}{IJCAI}, pp. \bibinfo{pages}{823--829}.
%Type = Inproceedings
\bibitem[{Haffari et~al.(2009)Haffari, Roy and
  Sarkar}]{haffari-etal-2009-active}
\bibinfo{author}{Haffari, G.}, \bibinfo{author}{Roy, M.},
  \bibinfo{author}{Sarkar, A.}, \bibinfo{year}{2009}.
\newblock \bibinfo{title}{Active learning for statistical phrase-based machine
  translation}, in: \bibinfo{booktitle}{Proceedings of Human Language
  Technologies: The 2009 Annual Conference of the North {A}merican Chapter of
  the Association for Computational Linguistics},
  \bibinfo{publisher}{Association for Computational Linguistics},
  \bibinfo{address}{Boulder, Colorado}. pp. \bibinfo{pages}{415--423}.
\newblock \URLprefix \url{https://aclanthology.org/N09-1047}.
%Type = Inproceedings
\bibitem[{Hieber et~al.(2018)Hieber, Domhan, Denkowski, Vilar, Sokolov, Clifton
  and Post}]{hieber-etal-2018-sockeye}
\bibinfo{author}{Hieber, F.}, \bibinfo{author}{Domhan, T.},
  \bibinfo{author}{Denkowski, M.}, \bibinfo{author}{Vilar, D.},
  \bibinfo{author}{Sokolov, A.}, \bibinfo{author}{Clifton, A.},
  \bibinfo{author}{Post, M.}, \bibinfo{year}{2018}.
\newblock \bibinfo{title}{The sockeye neural machine translation toolkit at
  {AMTA} 2018}, in: \bibinfo{booktitle}{Proceedings of the 13th Conference of
  the Association for Machine Translation in the {A}mericas (Volume 1: Research
  Track)}, \bibinfo{publisher}{Association for Machine Translation in the
  Americas}, \bibinfo{address}{Boston, MA}. pp. \bibinfo{pages}{200--207}.
\newblock \URLprefix \url{https://aclanthology.org/W18-1820}.
%Type = Article
\bibitem[{Hutchins and Lovtskii(2000)}]{Hutchins-2000}
\bibinfo{author}{Hutchins, J.}, \bibinfo{author}{Lovtskii, E.},
  \bibinfo{year}{2000}.
\newblock \bibinfo{title}{Petr petrovich troyanskii (1894–1950): A forgotten
  pioneer of mechanical translation}.
\newblock \bibinfo{journal}{Machine Translation} \bibinfo{volume}{15},
  \bibinfo{pages}{187–221}.
\newblock \URLprefix \url{https://doi.org/10.1023/A:1011653602669},
  \DOIprefix\doi{10.1023/A:1011653602669}.
%Type = Article
\bibitem[{Jelinek et~al.(1977)Jelinek, Mercer, Bahl and
  Baker}]{jelinek1977perplexity}
\bibinfo{author}{Jelinek, F.}, \bibinfo{author}{Mercer, R.L.},
  \bibinfo{author}{Bahl, L.R.}, \bibinfo{author}{Baker, J.K.},
  \bibinfo{year}{1977}.
\newblock \bibinfo{title}{Perplexity—a measure of the difficulty of speech
  recognition tasks}.
\newblock \bibinfo{journal}{The Journal of the Acoustical Society of America}
  \bibinfo{volume}{62}, \bibinfo{pages}{S63--S63}.
%Type = Inproceedings
\bibitem[{Kalchbrenner and Blunsom(2013)}]{kalchbrenner-blunsom-2013-recurrent}
\bibinfo{author}{Kalchbrenner, N.}, \bibinfo{author}{Blunsom, P.},
  \bibinfo{year}{2013}.
\newblock \bibinfo{title}{Recurrent continuous translation models}, in:
  \bibinfo{booktitle}{Proceedings of the 2013 Conference on Empirical Methods
  in Natural Language Processing}, \bibinfo{publisher}{Association for
  Computational Linguistics}, \bibinfo{address}{Seattle, Washington, USA}. pp.
  \bibinfo{pages}{1700--1709}.
\newblock \URLprefix \url{https://aclanthology.org/D13-1176}.
%Type = Inproceedings
\bibitem[{Kim et~al.(2006)Kim, Song, Kim, Cha and Lee}]{kim-etal-2006-mmr}
\bibinfo{author}{Kim, S.}, \bibinfo{author}{Song, Y.}, \bibinfo{author}{Kim,
  K.}, \bibinfo{author}{Cha, J.W.}, \bibinfo{author}{Lee, G.G.},
  \bibinfo{year}{2006}.
\newblock \bibinfo{title}{{MMR}-based active machine learning for bio named
  entity recognition}, in: \bibinfo{booktitle}{Proceedings of the Human
  Language Technology Conference of the {NAACL}, Companion Volume: Short
  Papers}, \bibinfo{publisher}{Association for Computational Linguistics},
  \bibinfo{address}{New York City, USA}. pp. \bibinfo{pages}{69--72}.
\newblock \URLprefix \url{https://aclanthology.org/N06-2018}.
%Type = Article
\bibitem[{Kingma and Ba(2014)}]{kingma2014adam}
\bibinfo{author}{Kingma, D.P.}, \bibinfo{author}{Ba, J.}, \bibinfo{year}{2014}.
\newblock \bibinfo{title}{Adam: A method for stochastic optimization}.
\newblock \bibinfo{journal}{arXiv preprint arXiv:1412.6980} .
%Type = Inproceedings
\bibitem[{Koehn et~al.(2007)Koehn, Hoang, Birch, Callison-Burch, Federico,
  Bertoldi, Cowan, Shen, Moran, Zens et~al.}]{koehn2007moses}
\bibinfo{author}{Koehn, P.}, \bibinfo{author}{Hoang, H.},
  \bibinfo{author}{Birch, A.}, \bibinfo{author}{Callison-Burch, C.},
  \bibinfo{author}{Federico, M.}, \bibinfo{author}{Bertoldi, N.},
  \bibinfo{author}{Cowan, B.}, \bibinfo{author}{Shen, W.},
  \bibinfo{author}{Moran, C.}, \bibinfo{author}{Zens, R.}, et~al.,
  \bibinfo{year}{2007}.
\newblock \bibinfo{title}{Moses: Open source toolkit for statistical machine
  translation}, in: \bibinfo{booktitle}{Proceedings of the 45th annual meeting
  of the association for computational linguistics companion volume proceedings
  of the demo and poster sessions}, pp. \bibinfo{pages}{177--180}.
%Type = Article
\bibitem[{Kreutzer et~al.(2019)Kreutzer, Bastings and
  Riezler}]{arxiv.1907.12484}
\bibinfo{author}{Kreutzer, J.}, \bibinfo{author}{Bastings, J.},
  \bibinfo{author}{Riezler, S.}, \bibinfo{year}{2019}.
\newblock \bibinfo{title}{Joey nmt: A minimalist nmt toolkit for novices}
  \URLprefix \url{https://arxiv.org/abs/1907.12484},
  \DOIprefix\doi{10.48550/ARXIV.1907.12484}.
%Type = Inproceedings
\bibitem[{Kunchukuttan et~al.(2018)Kunchukuttan, Mehta and
  Bhattacharyya}]{kunchukuttan-etal-2018-iit}
\bibinfo{author}{Kunchukuttan, A.}, \bibinfo{author}{Mehta, P.},
  \bibinfo{author}{Bhattacharyya, P.}, \bibinfo{year}{2018}.
\newblock \bibinfo{title}{The {IIT} {B}ombay {E}nglish-{H}indi parallel
  corpus}, in: \bibinfo{booktitle}{Proceedings of the Eleventh International
  Conference on Language Resources and Evaluation ({LREC} 2018)},
  \bibinfo{publisher}{European Language Resources Association (ELRA)},
  \bibinfo{address}{Miyazaki, Japan}.
\newblock \URLprefix \url{https://aclanthology.org/L18-1548}.
%Type = Inproceedings
\bibitem[{Laskar et~al.(2019)Laskar, Dutta, Pakray and
  Bandyopadhyay}]{inproceedings}
\bibinfo{author}{Laskar, S.}, \bibinfo{author}{Dutta, A.},
  \bibinfo{author}{Pakray, D.P.}, \bibinfo{author}{Bandyopadhyay, S.},
  \bibinfo{year}{2019}.
\newblock \bibinfo{title}{Neural machine translation: English to hindi}, pp.
  \bibinfo{pages}{1--6}.
\newblock \DOIprefix\doi{10.1109/CICT48419.2019.9066238}.
%Type = Inproceedings
\bibitem[{Liere and Tadepalli(1997)}]{Liere1997ActiveLW}
\bibinfo{author}{Liere, R.}, \bibinfo{author}{Tadepalli, P.},
  \bibinfo{year}{1997}.
\newblock \bibinfo{title}{Active learning with committees for text
  categorization}, in: \bibinfo{booktitle}{AAAI/IAAI}.
%Type = Inproceedings
\bibitem[{Liu et~al.(2018)Liu, Buntine and Haffari}]{Liu2018}
\bibinfo{author}{Liu, M.}, \bibinfo{author}{Buntine, W.},
  \bibinfo{author}{Haffari, G.}, \bibinfo{year}{2018}.
\newblock \bibinfo{title}{Learning to actively learn neural machine
  translation}, pp. \bibinfo{pages}{334--344}.
\newblock \DOIprefix\doi{10.18653/v1/K18-1033}.
%Type = Inproceedings
\bibitem[{Locke and Booth(1955)}]{Locke1955MachineTO}
\bibinfo{author}{Locke, W.N.}, \bibinfo{author}{Booth, A.D.},
  \bibinfo{year}{1955}.
\newblock \bibinfo{title}{Machine translation of languages: Fourteen essays}.
%Type = Inproceedings
\bibitem[{Luong et~al.(2015)Luong, Pham and
  Manning}]{luong-etal-2015-effective}
\bibinfo{author}{Luong, T.}, \bibinfo{author}{Pham, H.},
  \bibinfo{author}{Manning, C.D.}, \bibinfo{year}{2015}.
\newblock \bibinfo{title}{Effective approaches to attention-based neural
  machine translation}, in: \bibinfo{booktitle}{Proceedings of the 2015
  Conference on Empirical Methods in Natural Language Processing},
  \bibinfo{publisher}{Association for Computational Linguistics},
  \bibinfo{address}{Lisbon, Portugal}. pp. \bibinfo{pages}{1412--1421}.
\newblock \URLprefix \url{https://aclanthology.org/D15-1166},
  \DOIprefix\doi{10.18653/v1/D15-1166}.
%Type = Inproceedings
\bibitem[{Nguyen et~al.(2017)Nguyen, Daum{\'e}~III and
  Boyd-Graber}]{nguyen-etal-2017-reinforcement}
\bibinfo{author}{Nguyen, K.}, \bibinfo{author}{Daum{\'e}~III, H.},
  \bibinfo{author}{Boyd-Graber, J.}, \bibinfo{year}{2017}.
\newblock \bibinfo{title}{Reinforcement learning for bandit neural machine
  translation with simulated human feedback}, in:
  \bibinfo{booktitle}{Proceedings of the 2017 Conference on Empirical Methods
  in Natural Language Processing}, \bibinfo{publisher}{Association for
  Computational Linguistics}, \bibinfo{address}{Copenhagen, Denmark}. pp.
  \bibinfo{pages}{1464--1474}.
\newblock \URLprefix \url{https://aclanthology.org/D17-1153},
  \DOIprefix\doi{10.18653/v1/D17-1153}.
%Type = Article
\bibitem[{Olsson(2009)}]{Olsson2009}
\bibinfo{author}{Olsson, F.}, \bibinfo{year}{2009}.
\newblock \bibinfo{title}{A literature survey of active machine learning in the
  context of natural language processing} .
%Type = Inproceedings
\bibitem[{Papineni et~al.(2002)Papineni, Roukos, Ward and
  Zhu}]{papineni2002bleu}
\bibinfo{author}{Papineni, K.}, \bibinfo{author}{Roukos, S.},
  \bibinfo{author}{Ward, T.}, \bibinfo{author}{Zhu, W.J.},
  \bibinfo{year}{2002}.
\newblock \bibinfo{title}{Bleu: a method for automatic evaluation of machine
  translation}, in: \bibinfo{booktitle}{Proceedings of the 40th annual meeting
  of the Association for Computational Linguistics}, pp.
  \bibinfo{pages}{311--318}.
%Type = Inproceedings
\bibitem[{Peris and Casacuberta(2018)}]{peris-casacuberta-2018-active}
\bibinfo{author}{Peris, {\'A}.}, \bibinfo{author}{Casacuberta, F.},
  \bibinfo{year}{2018}.
\newblock \bibinfo{title}{Active learning for interactive neural machine
  translation of data streams}, in: \bibinfo{booktitle}{Proceedings of the 22nd
  Conference on Computational Natural Language Learning},
  \bibinfo{publisher}{Association for Computational Linguistics},
  \bibinfo{address}{Brussels, Belgium}. pp. \bibinfo{pages}{151--160}.
\newblock \URLprefix \url{https://aclanthology.org/K18-1015},
  \DOIprefix\doi{10.18653/v1/K18-1015}.
%Type = Article
\bibitem[{Post(2018)}]{arxiv.1804.08771}
\bibinfo{author}{Post, M.}, \bibinfo{year}{2018}.
\newblock \bibinfo{title}{A call for clarity in reporting bleu scores}
  \URLprefix \url{https://arxiv.org/abs/1804.08771},
  \DOIprefix\doi{10.48550/ARXIV.1804.08771}.
%Type = Misc
\bibitem[{Press and Wolf(2016)}]{arxiv.1608.05859}
\bibinfo{author}{Press, O.}, \bibinfo{author}{Wolf, L.}, \bibinfo{year}{2016}.
\newblock \bibinfo{title}{Using the output embedding to improve language
  models}.
\newblock \URLprefix \url{https://arxiv.org/abs/1608.05859},
  \DOIprefix\doi{10.48550/ARXIV.1608.05859}.
%Type = Inproceedings
\bibitem[{Ramanathan et~al.(2011)Ramanathan, Bhattacharyya, Visweswariah, Ladha
  and Gandhe}]{ramanathan-etal-2011-clause}
\bibinfo{author}{Ramanathan, A.}, \bibinfo{author}{Bhattacharyya, P.},
  \bibinfo{author}{Visweswariah, K.}, \bibinfo{author}{Ladha, K.},
  \bibinfo{author}{Gandhe, A.}, \bibinfo{year}{2011}.
\newblock \bibinfo{title}{Clause-based reordering constraints to improve
  statistical machine translation}, in: \bibinfo{booktitle}{Proceedings of 5th
  International Joint Conference on Natural Language Processing},
  \bibinfo{publisher}{Asian Federation of Natural Language Processing},
  \bibinfo{address}{Chiang Mai, Thailand}. pp. \bibinfo{pages}{1351--1355}.
\newblock \URLprefix \url{https://aclanthology.org/I11-1152}.
%Type = Inproceedings
\bibitem[{Rush(2018)}]{rush-2018-annotated}
\bibinfo{author}{Rush, A.}, \bibinfo{year}{2018}.
\newblock \bibinfo{title}{The annotated transformer}, in:
  \bibinfo{booktitle}{Proceedings of Workshop for {NLP} Open Source Software
  ({NLP}-{OSS})}, \bibinfo{publisher}{Association for Computational
  Linguistics}, \bibinfo{address}{Melbourne, Australia}. pp.
  \bibinfo{pages}{52--60}.
\newblock \URLprefix \url{https://aclanthology.org/W18-2509},
  \DOIprefix\doi{10.18653/v1/W18-2509}.
%Type = Article
\bibitem[{S(2017)}]{S2017StatisticalVR}
\bibinfo{author}{S, S.}, \bibinfo{year}{2017}.
\newblock \bibinfo{title}{Statistical vs rule based machine translation; a case
  study on indian language perspective}.
\newblock \bibinfo{journal}{ArXiv} \bibinfo{volume}{abs/1708.04559}.
%Type = Inproceedings
\bibitem[{Scheffer et~al.(2001)Scheffer, Decomain and
  Wrobel}]{Scheffer2001ActiveHM}
\bibinfo{author}{Scheffer, T.}, \bibinfo{author}{Decomain, C.},
  \bibinfo{author}{Wrobel, S.}, \bibinfo{year}{2001}.
\newblock \bibinfo{title}{Active hidden markov models for information
  extraction}, in: \bibinfo{booktitle}{IDA}.
%Type = Inproceedings
\bibitem[{Sennrich et~al.(2016)Sennrich, Haddow and
  Birch}]{sennrich-etal-2016-improving}
\bibinfo{author}{Sennrich, R.}, \bibinfo{author}{Haddow, B.},
  \bibinfo{author}{Birch, A.}, \bibinfo{year}{2016}.
\newblock \bibinfo{title}{Improving neural machine translation models with
  monolingual data}, in: \bibinfo{booktitle}{Proceedings of the 54th Annual
  Meeting of the Association for Computational Linguistics (Volume 1: Long
  Papers)}, \bibinfo{publisher}{Association for Computational Linguistics},
  \bibinfo{address}{Berlin, Germany}. pp. \bibinfo{pages}{86--96}.
\newblock \URLprefix \url{https://aclanthology.org/P16-1009},
  \DOIprefix\doi{10.18653/v1/P16-1009}.
%Type = Inproceedings
\bibitem[{Settles and Craven(2008a)}]{settles2008analysis}
\bibinfo{author}{Settles, B.}, \bibinfo{author}{Craven, M.},
  \bibinfo{year}{2008}a.
\newblock \bibinfo{title}{An analysis of active learning strategies for
  sequence labeling tasks}, in: \bibinfo{booktitle}{proceedings of the 2008
  conference on empirical methods in natural language processing}, pp.
  \bibinfo{pages}{1070--1079}.
%Type = Inproceedings
\bibitem[{Settles and Craven(2008b)}]{settles-craven-2008-analysis}
\bibinfo{author}{Settles, B.}, \bibinfo{author}{Craven, M.},
  \bibinfo{year}{2008}b.
\newblock \bibinfo{title}{An analysis of active learning strategies for
  sequence labeling tasks}, in: \bibinfo{booktitle}{Proceedings of the 2008
  Conference on Empirical Methods in Natural Language Processing},
  \bibinfo{publisher}{Association for Computational Linguistics},
  \bibinfo{address}{Honolulu, Hawaii}. pp. \bibinfo{pages}{1070--1079}.
\newblock \URLprefix \url{https://aclanthology.org/D08-1112}.
%Type = Inproceedings
\bibitem[{Seung et~al.(1992)Seung, Opper and Sompolinsky}]{sos-qc-92}
\bibinfo{author}{Seung, H.S.}, \bibinfo{author}{Opper, M.},
  \bibinfo{author}{Sompolinsky, H.}, \bibinfo{year}{1992}.
\newblock \bibinfo{title}{Query by committee}, in: \bibinfo{booktitle}{Proc.
  5th Annu. Workshop on Comput. Learning Theory}, \bibinfo{publisher}{ACM
  Press}. pp. \bibinfo{pages}{287--294}.
%Type = Misc
\bibitem[{Shen et~al.(2017)Shen, Yun, Lipton, Kronrod and
  Anandkumar}]{arxiv.1707.05928}
\bibinfo{author}{Shen, Y.}, \bibinfo{author}{Yun, H.}, \bibinfo{author}{Lipton,
  Z.C.}, \bibinfo{author}{Kronrod, Y.}, \bibinfo{author}{Anandkumar, A.},
  \bibinfo{year}{2017}.
\newblock \bibinfo{title}{Deep active learning for named entity recognition}.
\newblock \URLprefix \url{https://arxiv.org/abs/1707.05928},
  \DOIprefix\doi{10.48550/ARXIV.1707.05928}.
%Type = Misc
\bibitem[{Sutskever et~al.(2014)Sutskever, Vinyals and Le}]{arxiv.1409.3215}
\bibinfo{author}{Sutskever, I.}, \bibinfo{author}{Vinyals, O.},
  \bibinfo{author}{Le, Q.V.}, \bibinfo{year}{2014}.
\newblock \bibinfo{title}{Sequence to sequence learning with neural networks}.
\newblock \URLprefix \url{https://arxiv.org/abs/1409.3215},
  \DOIprefix\doi{10.48550/ARXIV.1409.3215}.
%Type = Inproceedings
\bibitem[{Tiedemann(2012)}]{tiedemann-2012-parallel}
\bibinfo{author}{Tiedemann, J.}, \bibinfo{year}{2012}.
\newblock \bibinfo{title}{Parallel data, tools and interfaces in {OPUS}}, in:
  \bibinfo{booktitle}{Proceedings of the Eighth International Conference on
  Language Resources and Evaluation ({LREC}'12)}, \bibinfo{publisher}{European
  Language Resources Association (ELRA)}, \bibinfo{address}{Istanbul, Turkey}.
  pp. \bibinfo{pages}{2214--2218}.
\newblock \URLprefix
  \url{http://www.lrec-conf.org/proceedings/lrec2012/pdf/463_Paper.pdf}.
%Type = Article
\bibitem[{Tur et~al.(2005)Tur, Hakkani-T{\"u}r and Schapire}]{tur2005combining}
\bibinfo{author}{Tur, G.}, \bibinfo{author}{Hakkani-T{\"u}r, D.},
  \bibinfo{author}{Schapire, R.E.}, \bibinfo{year}{2005}.
\newblock \bibinfo{title}{Combining active and semi-supervised learning for
  spoken language understanding}.
\newblock \bibinfo{journal}{Speech Communication} \bibinfo{volume}{45},
  \bibinfo{pages}{171--186}.
%Type = Misc
\bibitem[{Vaswani et~al.(2017)Vaswani, Shazeer, Parmar, Uszkoreit, Jones,
  Gomez, Kaiser and Polosukhin}]{arxiv.1706.03762}
\bibinfo{author}{Vaswani, A.}, \bibinfo{author}{Shazeer, N.},
  \bibinfo{author}{Parmar, N.}, \bibinfo{author}{Uszkoreit, J.},
  \bibinfo{author}{Jones, L.}, \bibinfo{author}{Gomez, A.N.},
  \bibinfo{author}{Kaiser, L.}, \bibinfo{author}{Polosukhin, I.},
  \bibinfo{year}{2017}.
\newblock \bibinfo{title}{Attention is all you need}.
\newblock \URLprefix \url{https://arxiv.org/abs/1706.03762},
  \DOIprefix\doi{10.48550/ARXIV.1706.03762}.
%Type = Inproceedings
\bibitem[{Wiseman and Rush(2016)}]{wiseman-rush-2016-sequence}
\bibinfo{author}{Wiseman, S.}, \bibinfo{author}{Rush, A.M.},
  \bibinfo{year}{2016}.
\newblock \bibinfo{title}{Sequence-to-sequence learning as beam-search
  optimization}, in: \bibinfo{booktitle}{Proceedings of the 2016 Conference on
  Empirical Methods in Natural Language Processing},
  \bibinfo{publisher}{Association for Computational Linguistics},
  \bibinfo{address}{Austin, Texas}. pp. \bibinfo{pages}{1296--1306}.
\newblock \URLprefix \url{https://aclanthology.org/D16-1137},
  \DOIprefix\doi{10.18653/v1/D16-1137}.
%Type = Inproceedings
\bibitem[{Zeng et~al.(2019)Zeng, Garg, Chatterjee, Nallasamy and
  Paulik}]{zeng-etal-2019-empirical}
\bibinfo{author}{Zeng, X.}, \bibinfo{author}{Garg, S.},
  \bibinfo{author}{Chatterjee, R.}, \bibinfo{author}{Nallasamy, U.},
  \bibinfo{author}{Paulik, M.}, \bibinfo{year}{2019}.
\newblock \bibinfo{title}{Empirical evaluation of active learning techniques
  for neural {MT}}, in: \bibinfo{booktitle}{Proceedings of the 2nd Workshop on
  Deep Learning Approaches for Low-Resource NLP (DeepLo 2019)},
  \bibinfo{publisher}{Association for Computational Linguistics},
  \bibinfo{address}{Hong Kong, China}. pp. \bibinfo{pages}{84--93}.
\newblock \URLprefix \url{https://aclanthology.org/D19-6110},
  \DOIprefix\doi{10.18653/v1/D19-6110}.
%Type = Inproceedings
\bibitem[{Zhang et~al.(2018)Zhang, Xu and Xiong}]{8629116}
\bibinfo{author}{Zhang, P.}, \bibinfo{author}{Xu, X.}, \bibinfo{author}{Xiong,
  D.}, \bibinfo{year}{2018}.
\newblock \bibinfo{title}{Active learning for neural machine translation}, in:
  \bibinfo{booktitle}{2018 International Conference on Asian Language
  Processing (IALP)}, pp. \bibinfo{pages}{153--158}.
\newblock \DOIprefix\doi{10.1109/IALP.2018.8629116}.
%Type = Inproceedings
\bibitem[{Zhao et~al.(2020)Zhao, Zhang, Zhou and Zhang}]{zhao-etal-2020-active}
\bibinfo{author}{Zhao, Y.}, \bibinfo{author}{Zhang, H.}, \bibinfo{author}{Zhou,
  S.}, \bibinfo{author}{Zhang, Z.}, \bibinfo{year}{2020}.
\newblock \bibinfo{title}{Active learning approaches to enhancing neural
  machine translation}, in: \bibinfo{booktitle}{Findings of the Association for
  Computational Linguistics: EMNLP 2020}, \bibinfo{publisher}{Association for
  Computational Linguistics}, \bibinfo{address}{Online}. pp.
  \bibinfo{pages}{1796--1806}.
\newblock \URLprefix \url{https://aclanthology.org/2020.findings-emnlp.162},
  \DOIprefix\doi{10.18653/v1/2020.findings-emnlp.162}.

\end{thebibliography}
%\end{thebibliography}
%\begin{biography}[{\includegraphics[width=1in,height=1.25in,clip,keepaspectratio]{mshell}}]{Michael Shell}

% ==== SWITCH OFF the BIO for submission
% ==== SWITCH OFF the BIO for submission

\end{document}